%% file: template.tex
\documentclass[a4paper]{article}

\usepackage{INTERSPEECH2018}
\usepackage{blindtext,color,tabularx,url}

\title{Learning to adapt: a meta-learning approach for speaker adaptation}

\name{Ond\v{r}ej Klejch, Joachim Fainberg, Peter Bell}
\address{
  Centre for Speech Technology Research, University of Edinburgh, Edinburgh EH8 9AB, UK}
\email{\{o.klejch, j.fainberg, peter.bell\}@ed.ac.uk}

\begin{document}

\maketitle
\begin{abstract}

The performance of automatic speech recognition systems can be improved by adapting an acoustic model to compensate for the mismatch between training and testing conditions, for example by adapting to unseen speakers. The success of speaker adaptation methods relies on selecting weights that are suitable for adaptation and using good adaptation schedules to update these weights in order not to overfit to the adaptation data. In this paper we investigate a principled way of adapting all the weights of the acoustic model using a meta-learning. We show that the meta-learner can learn to perform supervised and unsupervised speaker adaptation and that it outperforms a strong baseline adapting LHUC parameters when adapting a DNN AM with 1.5M parameters. We also report initial experiments on adapting TDNN AMs, where the meta-learner achieves comparable performance with LHUC.

\end{abstract}

\section{Introduction}

The accuracy of automatic speech recognition (ASR) systems can be improved by adapting their acoustic models (AM) to new speakers with available adaptation data. There are three possible approaches to speaker adaptation: feature-space approaches estimate feature transforms to maximise the log-likelihood of the adaptation data \cite{leggetter1995maximum,gales1998maximum}; model-based approaches update the parameters of the AM \cite{swietojanski2014learning, xue2014singular, liao2013speaker, yu2013kl}; and hybrid approaches use auxiliary features to provide information about the speaker to the AM \cite{dehak2011front,abdel2013fast}. 

In this paper we focus on model-based adaptation of neural network AMs. The aim is to compensate for a mismatch between training and testing data by adapting the weights of an AM. This may include adapting all the weights  \cite{liao2013speaker,huang2015regularized} or only a subset of the weights \cite{swietojanski2014learning,xue2014singular,yao2012adaptation,zhao2015investigating}. The biggest problem of adapting all the weights is that in scenarios with small amounts of adaptation data it tends to overfit. Methods that adapt only a subset of the weights try to overcome this problem by limiting the expressivity of the adaptation. Another problem is that adapting all the weights has a much bigger memory-footprint than adapting only a subset of the weights. 

Typically methods that adapt only a subset of the weights work by inserting new weights that scale activations of some hidden layer. For example, the Learning Hidden Unit Contributions (LHUC) method \cite{swietojanski2014learning} uses a speaker dependent vector $r$ (called LHUC parameters in the rest of the paper) as element-wise multipliers to scale the activations of a hidden layer $h$:

\begin{equation}
	h' = r \circ h.
\end{equation}
Other methods use a speaker dependent matrix $A$ to adapt the activations of a hidden layer $h$ \cite{gemello2007linear}:
\begin{equation}
	h' = Ah.
\end{equation}
Since the matrix $A$ can be quite big in terms of the number of speaker dependent parameters that have to be estimated using adaptation data, it is usually inserted after a layer with small dimensions. That can be a bottleneck layer or it can be inserted after a bottleneck obtained by decomposing a layer into a product of two low-rank matrices using singular value decomposition (SVD)~\cite{xue2014singular,zhao2017extended}.

Even adapting only a subset of the weights of a context-dependent AM might result in data sparsity issues arising from small amounts of adaptation data, as the posterior probabilities of senones that are unseen in the adaptation data are pushed towards zero~\cite{huang2015maximum}. Different regularisation techniques have been proposed to alleviate this issue, such as L2 normalisation of the difference between the original and the adapted weights~\cite{liao2013speaker}, Kullback-Leibler divergence of the posteriors of the original and the adapted model~\cite{yu2013kl} or maximum a posteriori (MAP) adaptation~\cite{huang2015maximum}. Alternatively, it has also been proposed to use a lower entropy task for adaptation, leveraging context-independent targets~\cite{swietojanski2015structured,huang2015rapid}.

While adapting only a subset of all weights benefits from a small memory footprint and a robustness to overfitting to adaptation data \cite{swietojanski2014learning}, it lacks the expressivity of adapting all weights. Therefore, we decided to find a reliable way of adapting all the weights of the AM which would eventually enable online and iterative adaptation~\cite{zhao2017extended}. To reliably adapt all the weights it is important to use a good adaptation schedule (number of adaptation steps, learning rate, etc.). Recently it has been shown that meta-learning can find good update rules for training neural networks \cite{andrychowicz2016learning,ravi2016optimization}, we therefore decided to evaluate it in the context of speaker adaptation. 

Meta-learning tries to replace hand-crafted algorithms with learned, specialised algorithms in a similar way to how deep learning replaced hand-crafted features with features extracted using neural networks. Recently, Andrychowicz et al.~\cite{andrychowicz2016learning} used a meta-learning approach to learn task specific update rules that outperform general update rules such as Adam~\cite{kingma2014adam}. They did so by training a coordinate-wise meta-learner that updates individual parameters of the trained model (see Section~2.1 for more details). The same approach was later used by Ravi and Larochelle~\cite{ravi2016optimization} to train a meta-learner for a few-shot learning scenario, where the meta-learner is required to train a new classifier for a set of unseen classes given only a few examples of each class (typically 1 or 5 examples per class). Speaker adaptation might be viewed as a special case of few-shot learning because we try to adapt a speaker-independent model to unseen speaker conditions using limited adaptation data. Therefore, we decided to evaluate the same meta-learning approach using the coordinate-wise meta-learner for speaker adaptation. There are several reasons for using meta-learning for speaker adaptation: 

\begin{enumerate}

  \item Speaker adaptation is performed many times, hence it is important to find a reliable adaptation schedule that will work for different speakers in different environments.
  \item The meta-learner learns how to adapt weights, but it ought to implicitly learn which sets of weights are suitable for adaptation in a given scenario.
  \item The meta-learner ought to find the best adaptation schedule given some external constraints such as the type of the AM, the number of adaptation steps, the amount of adaptation data, supervised/unsupervised targets, etc.

\end{enumerate}

\begin{figure*}[t]
  \includegraphics[width=\textwidth]{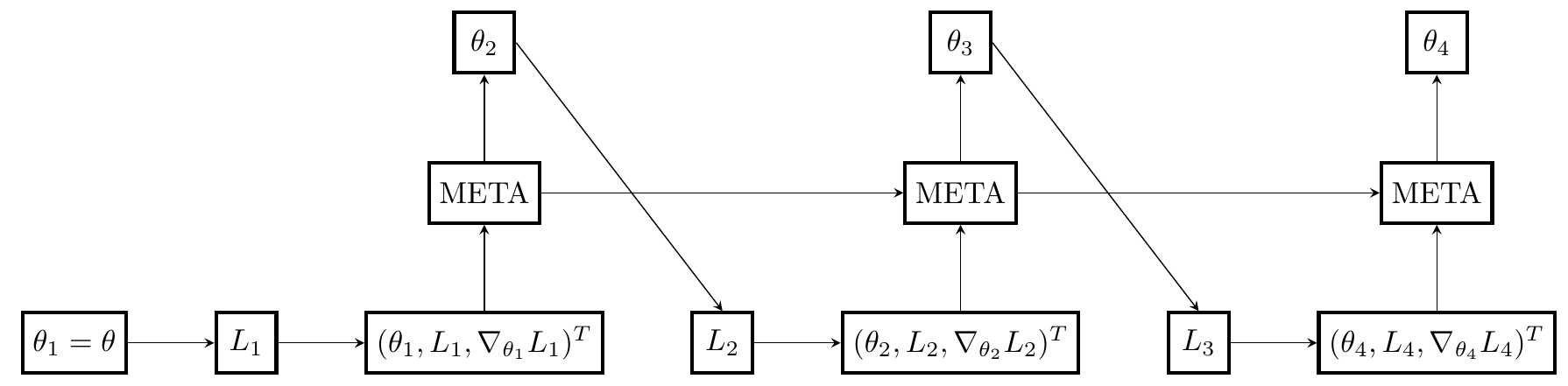}
  \caption{An illustration of how the meta-learner adapts a single weight $\theta$ in three adaptation steps.}
\end{figure*}

In this paper we investigate the use of a coordinate-wise meta-learner \cite{andrychowicz2016learning,ravi2016optimization} for supervised and unsupervised speaker adaptation and we compare it with adapting only LHUC parameters~\cite{swietojanski2014learning} and adapting all the weights of the AM. The rest of the paper is organised as follows. We formulate speaker adaptation as a meta-learning task and we describe the coordinate-wise meta-learner in Section~2, we describe adaptation experiments conducted in the domain of TED talks in Section~3 and we conclude the paper in Section~4.

\section{Method}

As stated in Section 1, meta-learning aims to replace handcrafted algorithms with learned task-specific algorithms. In order to train a meta-learner for a particular task it is necessary to find an appropriate meta-learning architecture and a loss function. In this section we describe these components for the meta-learner that we used for speaker adaptation.

\subsection{Speaker adaptation as a meta-learning task}

In this paper we view speaker adaptation as a function, $adapt$, with a set of parameters $\Phi$ that adapts a set of weights $\Theta$ of an acoustic model $f(x;\Theta)$ to a set of adapted weights $\Theta'$ using adaptation data $D=(X,Y)$, where $X$ are acoustic features and $Y$ are labels. It can be formalised by the equation

\begin{equation}
	adapt(f, \Theta, D; \Phi) \rightarrow \Theta'.
\end{equation}

Depending on the adaptation scenario, the labels $Y$ might correspond to true labels (supervised speaker adaptation) or labels obtained from a one-best path from a first pass decoding (unsupervised speaker adaptation). 

The meta-learner is then trained to adapt the weights of the original model using adaptation data such that it improves cross-entropy loss $L$ on test data. In this paper we approximate this by splitting the training data into $C$ equally sized chunks and training the meta-learner to adapt on chunk $c$ containing $N$ frames with corresponding data $D_{c} = (X_c, Y_c)$ to improve cross-entropy $L$ on the following chunk $c+1$ with data $D_{c+1} = (X_{c+1}, Y_{c+1})$. The loss $J$ for training of the meta-learner is

\begin{equation}
J(\Phi) = \sum_{c=1}^{C} \sum_{(x,y) \in D_{c+1}}{} L(y, f(x; adapt(f, \Theta, D_c, \Phi))).
\end{equation}

We do not need access to the test data for training the meta-learner. We can train the meta-learner with held-out data that was not used for training the AM, although it is beneficial if the held-out data resembles the test data as the meta-learner can learn the most suitable adaptation schedule for the AM for the given test set.

During evaluation the meta-learner uses adaptation data $D = (X, Y)$ to improve performance on test data by predicting a set of updated weights $\Theta'$ that are then used for decoding:

\begin{equation}
	\Theta' = adapt(f, \Theta, D; \Phi).
\end{equation}

It is important to note that in both the supervised and the unsupervised adaptation settings we use the true labels as $y_{test}$ for training the meta-learner, because we want to maximise performance on true labels. For $y_{adapt}$, however, we only use the true labels in the supervised speaker adaptation setting. In unsupervised speaker adaptation we use labels obtained from the first pass decoding.

\subsection{Coordinate-wise meta-learner}

The adaptation function, $adapt(f, \Theta, D; \Phi) \rightarrow \Theta'$, is implemented as a two layer LSTM with a set of parameters $\Phi$ as in \cite{ravi2016optimization}. The coordinate-wise meta-learner updates each weight $\theta \in \Theta$ individually -- each weight ~$\theta$ is presented as a single data sample to the meta-learner. Note, that in practice we batch all the weights $\Theta$ and we adapt them jointly. This has two advantages. First, the parameters of the meta-learner, $\Phi$, are shared across all weights $\Theta$ of the acoustic model. Second, the parameters of the meta-learner, $\Phi$, have much smaller dimensionality because they do not need to work with big inputs. In the following section we will describe how the meta-learner adapts a single weight $\theta \in \Theta$ using adaptation data $D$. The process is also illustrated in Figure~1.   

The first layer is a standard LSTM which at time step $t$ accepts a vector $v_t$ with three values: the current value of the weight $\theta_t$, the current loss value $L_t$, and the corresponding gradient $\nabla_{\theta_t} L_t$:
\begin{equation}
	v_t = \left(
    \begin{tabular}{c}
    $\theta_t$  \\
    $L_t$ \\
    $\nabla_{\theta_t} L_t$\\
    \end{tabular}
    \right).
\end{equation}
The loss $L_t$ is computed using adaptation data $D$ and current weights $\Theta_t$:
\begin{equation}
	L_t = \sum_{(x,y) \in D}{} L(y, f(x; \Theta_t)),
\end{equation}
and we initialise $\theta_1 = \theta$. Using this input vector $v_t$ the first LSTM layer produces a hidden representation
\begin{equation}
	h_t = LSTM(v_t).
\end{equation}
The second LSTM layer uses this hidden representation $h_t$ to predict the value of a forget gate $f_t$ with parameters $W_F$ and $b_F$:
\begin{equation}
	f_{t} = \sigma(W_F \cdot [h_t, f_{t-1}] + b_F),
\end{equation}
and to predict the value of an input gate $i_t$ with parameters $W_I$ and $b_I$:
\begin{equation}
    i_{t} = \sigma(W_I \cdot [h_t, i_t] + b_I).
\end{equation}
Both the forget gate $f_t$ and the input gate $i_t$ are used to update the weight $\theta_t$ to $\theta_{t+1}$ using the corresponding gradient $\nabla_{\theta_t} L$:
\begin{equation}
	\theta_{t+1} = f_{t} \cdot \theta_{t} - i_{t} \cdot \nabla_{\theta_t} L.
\end{equation}
In this update rule the input gate acts as a learning rate and the forget gate can be used to escape local minima when the loss is high but the gradient is close to zero~\cite{ravi2016optimization}.

Note that we followed \cite{andrychowicz2016learning} and preprocessed the losses $L_t$ with the following method using the suggested value $p = 10$: 

\begin{equation}
L_t \rightarrow 
\begin{cases}
    (\frac{log(|L_t|)}{p},sgn(x)) & \text{if } |L_t| \geq e^{p} \\
    (-1, e^{p}L_t), 	           & \text{otherwise.}
\end{cases}
\end{equation}
Similar preprocessing was applied to the gradients $\nabla_{\theta} L_t$.

\section{Experiments}

\subsection{Dataset}
All experiments were performed on TED talks. We used the TED-LIUM dataset \cite{rousseau2014enhancing} for training of the acoustic models and we tested these models on a combined test set of IWSLT 2010, 2011 and 2012 \cite{iwslt2010,iwslt2011,iwslt2012}.  To comply with the IWSLT evaluation protocol we only used talks that were recorded before the end of 2012 for training of the acoustic model. This pruned training set contains 134 hours of training data. The combined test set contains 30 speakers and is 5.3 hours long.

\subsection{Baseline DNN Setup}
Training the coordinate-wise meta-learners requires large amounts of memory -- linear in the number of weights of the acoustic model. Therefore, we decided to evaluate the adaptation methods on smaller acoustic models. We used Kaldi~\cite{povey2011kaldi} to train a small deep neural network model (called DNN in Table~1 and Table~2) with 1.5M weights across 6 hidden layers, each with 256 neurons, using sigmoid activation functions and an output layer corresponding to 3792 tied context-dependent phones and an input corresponding to 7 acoustic frames. 

We also experimented with the adaptation of a time-delayed neural network model (TDNN) \cite{peddinti2015time} to show that the meta-learner works with state-of-the-art architectures. Again we trained a smaller model with 2.1M weights across 6 hidden layers, with 300 units each and RELU activation function and splice indexes \texttt{-2,-1,0,1,2 -1,2 -3,3 -7,2 -3,3 0}. We evaluated the adaptation of two TDNN models, one using batch-normalisation and the other using L2 normalisation\footnote{\texttt{y = x * sqrt(dim(x)) * target-rms / $|$x$|$}} (TDNN-BN and TDNN-RN in Table~1 and Table~2). All models used a pruned 3-gram language model for decoding, we did not use a 4-gram language model for rescoring. 

\subsection{Adaptation Setup}
We explored adaptation using only a small amount of adaptation data: in all the experiments we used the first 10 seconds of data to perform speaker adaptation. For the baseline adaptation experiments we either adapted all layers or only the LHUC parameters of each unit in each layer (denoted ALL and LHUC in Table~1 and Table~2). For both technique we adapted for 3 epochs using stochastic gradient descent (SGD) with a learning rate of: $0.01$ for the DNN model; $2.5 \cdot 10^{-6}$ for the TDNN models (ALL); and $0.7$ (LHUC) for both DNN and TDNN models.

We used the development sets from IWSLT~2010 and 2012 \cite{iwslt2010,iwslt2012} to train the meta-learner. Together these datasets contain 3 hours of audio. The data for the first 13 speakers was used as the training set and the last 5 speakers as the validation set. We trained the meta-learner to adapt the acoustic model using 10~seconds of adaptation data to improve performance on the following 10 seconds. The meta-learner was trained using Adam \cite{kingma2014adam} with a learning rate of $0.001$. During training we monitored the loss on the validation set. We selected the meta-learner that achieved the best validation loss for testing. Note that we initialized a bias of the input gate $b_I$ to small values (sampled uniformly from $[-5, -4]$) and a bias of the forget gate $b_F$ to high values (sampled uniformly from $[4, 5]$) such that the meta-learner starts training with an update rule similar to SGD.

When training a meta-learner for the DNN model we split the data into chunks of 1000 frames and we trained a meta-learner with 20 units in the hidden layer to perform one full-batch adaptation step. We also experimented with more adaptation steps, but we did not observe any improvements compared to performing a single adaptation step. When we tried the same procedure with TDNN models, the meta-learner learned to overfit to silent frames, producing too many deletion errors. This was surprising as the amount of silent frames in the alignments obtained with TDNN models was similar to those obtained by the DNN model. We hypothesized that this might be due to the much larger context (29 acoustic frames) of the TDNN models, so we therefore removed silent phones from the meta-learner training data and from the adaptation data -- there is not much benefit from doing speaker adaptation on silent frames. We preprocessed the data by trimming silence at the beginning and end of each utterance, splitting each utterance into 50~ms chunks, and discarding chunks that contained more than $10\%$ silent frames.  Next we also changed the input vector to the meta-learner, $v$, to contain the normalised position of the weight in the weight matrix instead of the weight value $\theta$. If the weight $\theta$ corresponded to an element at a position $i, j$ in a weight matrix $W \in R^{m \times n}$ then the input vector $v$ was:
\begin{equation}
	v_t = \left(
    \begin{tabular}{c}
    $i / m$  \\
    $j / n$ \\
    $L_t$ \\
    $\nabla_{\theta_t} L_t$\\
    \end{tabular}
    \right).
\end{equation}
This should provide the meta-learner with much finer control over which weights are suitable for adaptation instead of just learning different adaptation schedules for each layer. 

We implemented the meta-learner\footnote{\url{https://github.com/choko/learning_to_adapt}} in Tensorflow \cite{abadi2016tensorflow} and Keras \cite{chollet2015keras}.

\subsection{Results}

In this paper we evaluated relatively small models in order to be able to train a meta-learner for them. The DNN model achieves a word error rate (WER) of $20.7\%$ on our test set and both TDNN models achieve a WER of $15.2\%$ which is close to a model used in \cite{bell2014uedin} which achieves WER $14.9\%$ while the TDNN models are $20\times$ smaller. 

\input{tables/supervised}
\input{tables/unsupervised}

First we performed supervised speaker adaptation using 10~seconds of adaptation data (Table~1). When adapting the LHUC parameters of the DNN model we got an improvement of $0.6\%$ absolute which is comparable to other previous experiments~\cite{swietojanski2014learning}. Adapting all the weights performed  worse as the adaptation schedule may have overfitted to the adaptation data, but the meta-learner was able to learn a good adaptation schedule that outperforms adapting LHUC parameters by another $0.3\%$ absolute. Note that the meta-learner performed only 1 adaptation step and LHUC used 3 adaptation steps. We believe that finding a way to train the meta-learner for more adaptation steps might bring further improvements. When adapting TDNN models the meta-learners were not that successful, they were not able to match performance of LHUC (TDNN-BN) or outperform LHUC (TDNN-RN). The meta-learner may require more tuning to be able work with such complicated models as TDNNs. It is interesting to point out that the TDNN with batch normalisation benefits much more from LHUC adaptation than the TDNN trained with L2 normalisation. This finding is similar to \cite{wang2017unsupervised} where they adapted only batch normalisation parameters during speaker adaptation.

We then performed unsupervised adaptation using 10~seconds of adaptation data (Table~2). When adapting the DNN model, the meta-learner outperforms adapting only LHUC parameters. To test whether the meta-learner is able to determine the best adaptation in the given scenario, we evaluated a meta-learner that was trained to use supervised targets for adaptation (META-sup) and a meta-learner that was trained to use unsupervised targets for adaptation (META-unsup). The results show that it is very beneficial to train the meta-learner in the same conditions as in testing. Unfortunately, unsupervised adaptation of TDNN models with a current implementation of the meta-learner did not match the performance of LHUC. 

\section{Conclusions}

In this paper we tried to find a reliable way of adapting all the weights of a neural network AM because we believe that adapting only a subset of the weights limits expressivity of the adaptation. We investigated a meta-learning approach for finding a good adaptation schedule -- we showed that speaker adaptation can be formulated as a meta-learning task by defining a loss to train a coordinate-wise meta-learner as in \cite{andrychowicz2016learning,ravi2016optimization} to perform supervised and unsupervised speaker adaptation using only 10~seconds of adaptation data. We compared this approach with adapting all weights of the AM and LHUC parameters using SGD. Our results on adaptation of the DNN model suggest that meta-learning might be a useful method for speaker adaptation, especially for unsupervised speaker adaptation when the meta-learner is trained to perform unsupervised speaker adaptation. So far we were not able to outperform adaptation of LHUC parameters in TDNN models but we believe that the meta-learner ought to learn better adaptation strategy when properly trained.

In future work we want to focus on better understanding the training of the meta-learner. In this paper we used around 3 hours of data divided between 18 speakers to train the meta-learner. It is possible that because of this small training dataset the meta-learner is overfitting to speakers in the training data. Therefore, we plan to train the meta-learner with much larger training datasets to enable it to learn inter-speaker variability.

Another issue in training of the meta-learner is that the loss, as formulated in the Equation~(4), does not penalise the meta-learner when it learns that the best strategy is to overfit to senones seen in the adaptation data. For example, when there is a big overlap in senones between two consecutive chunks the best strategy for the meta-learner is to predict only senones seen in the adaptation data. Therefore, we need to study how to regularise the meta-learner to not learn to overfit to adaptation data, either by using L1/L2 penalty on the difference of the adapted and the original weights~\cite{liao2013speaker} or by using Kullback-Leibler divergence between the posteriors of the adapted and the original model~\cite{yu2013kl}. In preliminary experiments we tried to use L1 regularisation which forced the meta-learner to learn to adapt only some layers, but it did not improve the accuracy of the adapted model. Alternatively, we could use a lower entropy task for adaptation~\cite{swietojanski2015structured, huang2015rapid}, however we believe that a more expressive form of the meta-learner ought to figure out a relation between senones and monophones and co-adapt senones that correspond to the same phone. 

Finally, it is necessary to explore ways of scaling the meta-learning both to larger models and to larger amounts of adaptation data. The problem with scaling to larger models is that it is not possible to fit a meta-learner to the GPU memory as the memory consumption is linear in the number of weights of an AM. Therefore, we would like to reparameterise the meta-learner to learn to perform some form of low-rank adaptation with lower memory requirements. Lower memory requirements would allow us to perform more adaptation steps without using truncated backpropagation through time.

\section{Acknowledgements}

This work was partially supported by the H2020 project SUMMA, under grant agreement 688139, and a PhD studentship funded by Bloomberg. We would like to thank Steve Renals for many insightful discussions.

\bibliographystyle{IEEEtran}
\bibliography{mybib}

\end{document}

%% file: tables/supervised.tex
\begin{table}[t!]
\begin{center}
\begin{tabularx}{\columnwidth}{Xccc}
						& 	DNN			&	TDNN-BN	&	TDNN-RN \\
\hline
original				&	20.7	 	&	15.2 	&	15.2 \\
LHUC					&	20.1		&	14.3	&	14.6 \\
ALL						&	20.6		&	14.5	&	14.6 \\
META					& 	19.8		&	14.4	&	14.6 \\
\hline
\end{tabularx}

\caption{WER (\%) of the supervised speaker adaptation experiments using 10s of adaptation data.}
\end{center}
\end{table}

%% file: tables/unsupervised.tex
\begin{table}[t!]
\begin{center}
\begin{tabularx}{\columnwidth}{Xccc}
						& 	DNN		&	TDNN-BN	&	TDNN-RN \\
\hline
original				&	20.7 	&	15.2 	&	15.2 \\
LHUC					&	20.5	&	14.6	&	14.8 \\
ALL						&	20.6	&	14.7	&	14.8 \\
META-sup				& 	20.0	&	14.7	&	15.0 \\
META-unsup				& 	19.7	&	14.7	&	14.9 \\
\hline
\end{tabularx}

\caption{WER (\%) of the unsupervised speaker adaptation experiments using 10s of adaptation data.}
\end{center}
\end{table}